\documentclass{bmvc2k}

\title{Learnable Gated Temporal Shift Module for \\ Deep Video Inpainting}

\addauthor{Ya-Liang Chang*}{yaliangchang@cmlab.csie.ntu.edu.tw}{1}
\addauthor{Zhe Yu Liu*}{zhe2325138@cmlab.csie.ntu.edu.tw}{1}
\addauthor{Kuan-Ying Lee}{r03922165@ntu.edu.tw}{1}
\addauthor{Winston Hsu}{whsu@ntu.edu.tw}{1}

\addinstitution{
National Taiwan University, \\
Taipei, Taiwan
}

\runninghead{Chang, Liu, Lee and Hsu}{LGTSM for Deep Video Inpainting}

\def\etal{\emph{et al}\bmvaOneDot}
\usepackage{amsmath}
\usepackage{amssymb}
\usepackage{dsfont}
\usepackage{array,multirow,graphicx}
\usepackage{pdfrender}
\newcommand*{\boldcheckmark}{%
  \textpdfrender{
    TextRenderingMode=FillStroke,
    LineWidth=.5pt, 
  }{\checkmark}%
}

  \usepackage{etoolbox,lipsum}
  \usepackage{multirow}
  
\begin{document}

\maketitle

\begin{abstract}
How to efficiently utilize temporal information to recover videos in a consistent way is the main issue for video inpainting problems. Conventional 2D CNNs have achieved good performance on image inpainting but often lead to temporally inconsistent results where frames will flicker when applied to videos (see \href{https://www.youtube.com/watch?v=87Vh1HDBjD0&list=PLPoVtv-xp_dL5uckIzz1PKwNjg1yI0I94&index=1}{video:Edge-Connect}); 3D CNNs can capture temporal information but are computationally intensive and hard to train. In this paper, we present a novel component termed Learnable Gated Temporal Shift Module (LGTSM) for video inpainting models that could effectively tackle arbitrary video masks without additional parameters from 3D convolutions. LGTSM is designed to let 2D convolutions make use of neighboring frames more efficiently, which is crucial for video inpainting. Specifically, in each layer, LGTSM learns to shift some channels to its temporal neighbors so that 2D convolutions could be enhanced to handle temporal information. Meanwhile, a gated convolution is applied to the layer to identify the masked areas that are poisoning for conventional convolutions. On the FaceForensics and Free-form Video Inpainting (FVI) dataset, our model achieves state-of-the-art results with simply 33\% of parameters and inference time. The source code is available on \url{https://github.com/amjltc295/Free-Form-Video-Inpainting}.
\end{abstract}

\begin{figure} [htb]
\centering
\includegraphics[width=\linewidth]{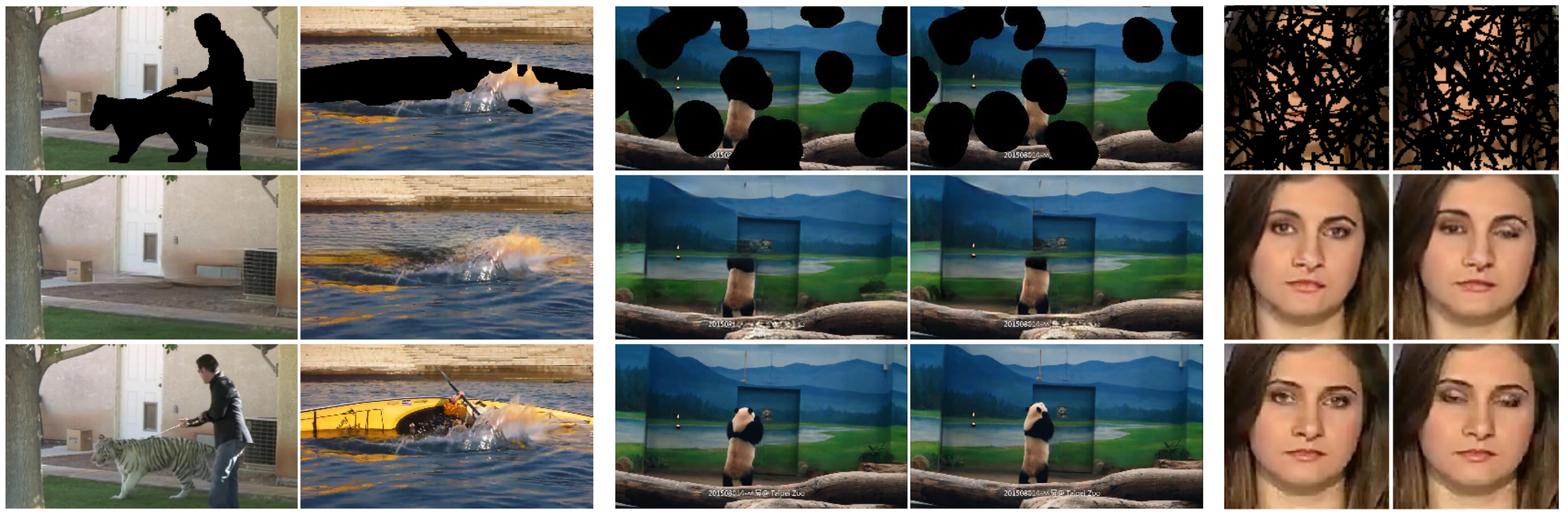}
\caption{Our model takes videos with free-form masks (first row) and fills in the missing areas with proposed LGTSM to generate realistic completed results (second row) compared to the original videos (third row). It could be applied to video editing tasks such as video object removal, as shown in the first two columns. Best viewed in color and zoom-in. See corresponding videos in the following links: \href{https://www.youtube.com/watch?v=585xZjcmUlA&list=PLPoVtv-xp_dIc_qhYe5lrOyWCLEu7hWoL&index=4&t=0s}{object removal}, \href{https://www.youtube.com/watch?v=08MNqZla29g&list=PLPoVtv-xp_dL5uckIzz1PKwNjg1yI0I94&index=37}{free-form masks}, and \href{https://www.youtube.com/watch?v=3uiOOyimBHw&list=PLPoVtv-xp_dKLIZkMBvhhCu97AsXQ9e_D&index=2&t=0s}{faces}.
}
\label{fig:teaser}
\end{figure}

\section{Introduction}
Free-form video inpainting, to fill in arbitrary missing regions in a video, is a very challenging task. It could be widely used for movie post-processing, damaged video recovery and video editing. For humans, it takes tremendous efforts to recover these missing areas like Fig. \ref{fig:teaser}, while an autonomous method may complete it easily.

The key for free-form video inpainting is to model spatial-temporal features. That is, a model needs to capture the content of masked areas according to its surroundings and fill in these areas with related pixels. Traditional patch-based methods \cite{Huang-SigAsia-2016,newson2014video} fill in these areas by finding similar patches from other parts of the videos. However, the searching algorithms usually have high computational complexity and the missing area may not be found for complex objects or masks (see Fig. \ref{fig:visual_comparison}).

On the other hand, deep learning methods \cite{yu2018free, nazeri2019edgeconnect, liu2018image, chang2019free, chang2019vornet} could fill in unseen masked areas by the encoding and decoding process, based on the structures learned from the training data. Still, compared to the success in image inpainting, deep learning methods struggle to model these video features due to the additional temporal dimension. Using 3D convolution to model spatial-temporal features is the most intuitive way but it requires plenty parameters and is hard to train.

In this paper, we propose a novel component termed Learnable Gated Temporal Shift Module (LGTSM) to handle free-form video masks with 2D convolutions, motivated by the TSM originally for action recognition. Though inspired by TSM, we found that TSM perfectly suitable for free-form video inpainting as it cannot totally make use of neighboring frames from the beginning layers nor handle irregular masks, which makes us propose LGTSM. Specifically, in each layer, LGTSM learns to shift a part of feature channels in a frame to its neighboring frame, and then attends on masked/inpainted/unmasked areas by a gating convolutional filter. LGTSM enables 2D convolutions to process masked videos and generate state-of-the-art results as 3D convolutions with only 33\% parameters and inference time.

Our paper makes the following contributions:
\begin{itemize}
\item We propose the Gated Temporal Shift Module that could recover videos with free-form masks in 2D convolutions by temporally shifting and gating features in each layer, which reduce the model size and computational time to 33\% compared to 3D convolutions.
\item Given that video inpainting requires more information from neighboring frames, we propose a novel Learnable Gated Temporal Shift Module that could learn the temporally shifting kernels and achieved state-of-the-art performance.
\item We propose the TSMGAN loss which significantly improves the model performance for free-form video inpainting.
\end{itemize}

\section{Related Work}
\paragraph{Image Inpainting.}

Recently, deep learning based methods have taken over the image inpainting task. Xie \etal \cite{xie2012image} firstly apply convolutional neural networks (CNNs) on small-region image inpainting and denoising. Pathak \etal \cite{pathak2016context} then extended \cite{xie2012image} to larger region with an encoder-decoder structure. Moreover, \cite{pathak2016context}  adopted the generative adversarial network (GAN) \cite{goodfellow2014generative}, where a generator learning to create realistic images and a discriminator striving to tell fake ones are trained together to improve the image quality. Subsequently, \cite{yu2018generative, yu2018free, nazeri2019edgeconnect} also developed new GAN architectures with different components for image inpainting. 

Among these deep methods, Yu \etal \cite{yu2018free} proposed gated convolutions for image inpainting that uses an additional gating convolution to learn the difference between masked, inpainted and unmasked areas in each layer. We integrate such gating mechanism to our model. Also, Nazeri \etal \cite{nazeri2019edgeconnect} developed a two-stage model that generates image edges first before recovering the whole images conditioned on edges. Their model achieved state-of-the-art results, and we set it as one of our baselines.

\vspace{-5mm}
\paragraph{Video Inpainting.}
Generally, video inpainting could be viewed as an extension of image inpainting with temporal constraints (i.e content in different frames need to be consistent.) However, different from image inpainting, patch-based methods \cite{granados2012not, Huang-SigAsia-2016, newson2014video, wexler2007space} still play a role in video inpainting as more patches are available in videos.  Among them, Huang \etal \cite{Huang-SigAsia-2016} jointly estimate optical flow and colors in the masked region to fix the moving camera problem and reached state-of-the-art results, so we also set it as one of our baselines.

Although patch-based methods have made great success in video inpainting, they are highly limited in computational time due to the search algorithms. In addition, the masked areas still need to be patchable; these methods do not work on complex objects such as faces. To address these problems, Wang \etal \cite{wang2018videoinp} proposed the first deep learning based method for video inpainting, with a two-stage CombCN that uses 3D convolutions to generate coarse but temporally consistent videos and then refines with 2D convolutions. Their model could learn to recover face videos so we also set it as a baseline.

\begin{figure} \centering
\includegraphics[width=\linewidth]{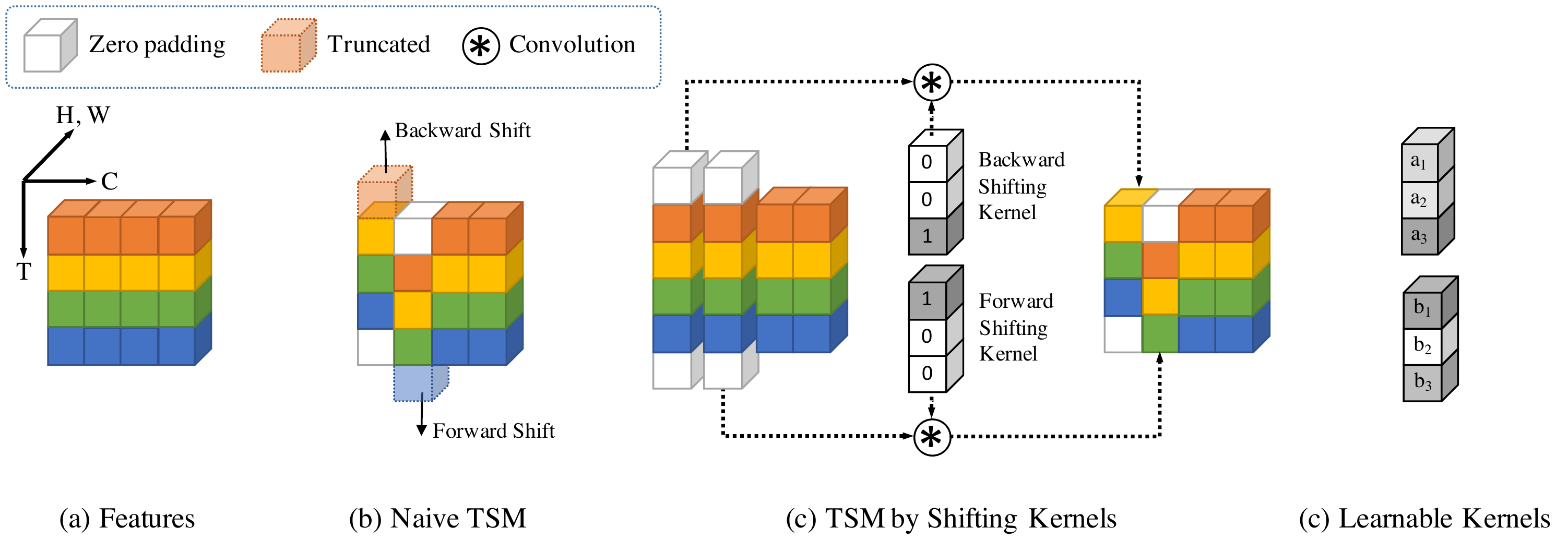}
\caption{Explanation of the learnable shifting kernels in the proposed LGTSM. (a) Input features for the layer. We will do shifting operation on channel $\times$ time dimensions. (b) Original TSM from \cite{lin2018temporal}. (c) Equivalent TSM by temporal shifting kernels. (d) In the proposed LGTSM, temporal shifting kernels are also learnable and the size could be different.}
\label{fig:LTSM}
\end{figure}

\vspace{-5mm}
\paragraph{Temporal Modeling.}
As most state-of-the-art deep video inpainting learning methods adopt the encoder-decoder structure, the key is to model the spatial-temporal structures in videos. Over the past few years, a variety of deep learning architectures have been proposed to model video structures, especially for action recognition. These architectures include applying temporal pooling \cite{karpathy2014large} or recurrent networks \cite{yue2015beyond} on top of 2D convolutions to model temporal features structures, combining optical flows and RGB frames to make two-stream networks \cite{simonyan2014two}, directly using 3D convolutions \cite{tran2015learning} and variants of these models \cite{feichtenhofer2016convolutional, carreira2017quo}. For more details, we refer readers to \cite{carreira2017quo}.

Despite the great performance, many architectures for action recognition cannot be applied to video inpainting since 1) the input video is corrupted, so it is hard to derive optical flows or apply naive convolutions 2) these architectures only need an encoder, while for video inpainting requires a decoder to recover the missing areas. 

Aside from architecture-level temporal modeling, there are also works that focus on the module-level \cite{lin2018temporal, li2018temporal}, which is more applicable for video inpainting.  Lin \etal \cite{lin2018temporal} proposed Temporal Shit Module that shifts part of feature channels in each frame to its neighboring frames so that 2D convolutions could handle temporal information. Similarly, Li \etal \cite{li2018temporal} developed Temporal Bilinear (TB) that applies factorized bilinear operation on features to model interactions between frames. Note that these models are for action recognition that all input frames are valid, while for video inpainting, many pixels are masked. Based on these ideas of integrating temporal information to 2D convolutions, we propose the Learnable Gated Temporal Shift Module for video inpainting.

\section{Proposed Method}
\subsection{Learnable Gated Temporal Shift Module}  \label{GatedTSM}
Models with 3D convolutions could capture temporal information in an intuitive way but are hard to train due to the large number of parameters. To this end, we extend the residual Temporal Shift Module (TSM) \cite{lin2018temporal}, originally designed for action recognition, to video inpainting. TSM tackles temporal information in 2D convolutions. The input activation $F_{t, x, y}$ for each convolutional layer in video inpainting is in shape of (B, C, L, H, W) where B is the batch size, C is the channel number, L is the temporal length, H is the height and W is the width of the input activation. For each frame in L, TSM shifts a portion of channels to its previous and next frame  before the convolutional operations, as shown in Fig. \ref{fig:LTSM}(b). These shifted channels contain features from other frames, so together with unshifted features, the original 2D convolutions could learn the temporal structures accordingly. 

\begin{figure} \centering
\includegraphics[width=\linewidth]{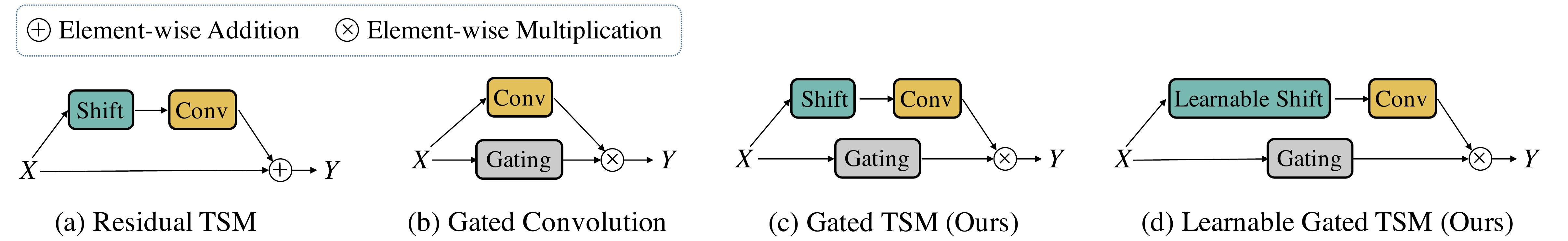}
\caption{Module design. We integrate (a) Residual TSM \cite{lin2018temporal} and (b) gated convolution \cite{yu2018free} to (c) Gated Temporal Shift Module (GTSM) and design learnable temporal shifting kernel (Fig. \ref{fig:LTSM}c) to make (d) the proposed Learnable Gated Temporal Shift Module (LGTSM).}
\label{fig:blocks}
\end{figure}

However, for free-form video inpainting, not every feature point is valid as many areas are masked. These masked areas are harmful to naive TSM as convolutions cannot tell the difference between valid and invalid feature points. To address this issue, we design the Gated Temporal Shift Module (GTSM) for free-form video inpainting (see Fig. \ref{fig:blocks}). Specifically, in addition to the TSM module, a gating convolutional filter $W_g$ is applied to input features $F_{t, x, y}$ to obtain a gating $Gating_{t, x, y}$. This gating will serve as a soft validity map to identify the masked/unmasked/inpainted areas for the output features $Features_{t, x, y}$ from the TSM module with original convolutional filter $W_f$. 

Mathematically, GTSM could be expressed as:

\begin{equation}
Gating_{t, x, y} = \sum \sum W_g \cdot F_{t, x, y} 
\end{equation}
\vspace{-2mm}
\begin{equation}
Features_{t, x, y} = \sum \sum W_f \cdot TSM(F_{t-1, x, y}, F_{t, x, y}, F_{t+1, x, y})
\end{equation}
\vspace{-2mm}
\begin{equation}
Output_{t, x, y} = \sigma(Gating_{t, x, y})  \phi(Features_{t, x, y})
\end{equation}
where $\sigma$ is the sigmoid function that transforms gating to values between 0 (invalid) and 1 (valid), and $\phi$ is the activation function for the convolution. Note that the TSM module could be easily modified to online settings (without peeking future frames) for real-time applications.

Note that the temporal shifting operation in TSM is similar to applying forward/backward shifting kernels on the channel-temporal map, as shown in Fig. \ref{fig:LTSM}(c). In TSM, these kernels are fixed; a frame could only get features from its one-frame neighbors in each layer. Such fixed shifting kernels in TSM are insufficient to make use of further neighboring frames as temporal information could only be aggregated through deeper layers.
Unlike action recognition, video inpainting models sorely need information from the beginning layers to capture the spatial-temporal structures so that deeper layers could recover the missing areas accordingly. 

Therefore, we also propose the Learnable Gated Temporal Shift Module (LGTSM), where the temporal shifting kernels are also learnable and the kernel size could be larger (see Fig.  \ref{fig:LTSM}(d)). With LGTSM, the model could learn to shift and scale features from specific temporal neighbors in each layer (or not). For example, the model could get features from more temporally further neighbors in the first few layers and remain unshifted in the deeper layers. This greatly enhances the model capability with very little cost.

In practice, 
the shifting operation only uses an additional buffer in size of 1/4 channels, so it has little cost in terms of computational time and run-time memory compared to traditional 2D convolutions. Note that the number of kernels could also be flexible (there are only two for TSM: forward and backward). Moreover, LGTSM could learn temporal information with very few extra parameters, and we found that it could achieve state-of-the-art results as the 3D convolutions with only 33\% parameters and inference time.
\subsection{Loss Functions}
Design the combination of loss functions to train a video inpainting model is non-trivial due to the uncertainty of the free-form masks and the high complexity of video features. We use $l_1$ loss for low-level features, perceptual loss and style loss for image content, and propose TSMGAN loss for handle high-level features and enhance realness. In this section, we will introduce these loss functions to train our model. 

\vspace{-3mm}
\paragraph{\textbf{Masked $l_1$ loss.}} The $l_1$ loss focuses on the pixel-level features, which is widely used for generative tasks on image \cite{yu2018generative,yu2018free,liu2018image,nazeri2019edgeconnect}, and videos \cite{wang2018videoinp,chang2019free}:
\begin{equation}
L_{l_1} = \mathds{E}_{t,x,y}[ |O_{t,x,y} - V_{t, x, y}|]
\end{equation}

\vspace{-3mm}
\paragraph{\textbf{Perceptual loss and style loss.}} $l_1$ loss often leads to blurry results as it only focus on low-level features. To address this problem, we adopt the perceptual and style loss \cite{gatys2015neural} to keep image contents. Similar loss functions could be found in many generative tasks such as image inpainting \cite{liu2018image, nazeri2019edgeconnect} and super-resolution \cite{johnson2016perceptual, ledig2017photo}. 

The perceptual loss could be viewed as the $l_1$ loss in feature level:
\begin{equation} \label{equ6}
L_{perc} = \sum_{t=1}^{n} \sum_{p=0}^{P-1} \frac{|\Psi^{O_{t}}_p - \Psi^{V_{t}}_p|}{N_{\Psi^{V_{t}}_p}}
\vspace{-2mm}
\end{equation}
where $V_t$ is the input video, $\Psi^{V_{t}}_p$ is the activation from the $p$th selected layer of the pretrained network, and $N_{\Psi^{V_{t}}_p}$ is the number of elements in the $p$th layer. We choose layer $relu_{2\_2}$, $relu_{3\_3}$ and $relu{4\_3}$ from the VGG \cite{simonyan2014very} network pre-trained on ImageNet \cite{russakovsky2015imagenet}.

Style loss is a variant of perceptual loss, with an auto-correlation (Gram matrix) applied to the features first:
\begin{equation}
L_{style} = \sum_{t=1}^{n} \sum_{p=0}^{P-1} \frac{1}{C_p C_p} \frac{|(\Psi^{{O_t}_p})^T(\Psi^{{O_t}_p}) - (\Psi^{V_{t}}_p)^T(\Psi^{V_{t}}_p))|}{C_p H_p W_p}
\end{equation}
where $\Psi^{O_{t}}_p$ and $\Psi^{{V_t}_p}$ are both features from the pre-trained VGG network, as the ones in the perceptual loss \ref{equ6}.

\vspace{-3mm}
\paragraph{\textbf{TSMGAN loss.}}
All the aforementioned loss functions are for image only, which do not take temporal consistency into consideration. Therefore, we develop TSMGAN to learn temporal consistency.  We set up a generative adversarial network (GAN) with Gated Temporal Shit Module integrated on both the generator and discriminator as stated in \ref{GatedTSM}. 

The TSMGAN discriminator is composed of six 2D convolutional layers with TSM. Also, we apply the recently proposed spectral normalization \cite{miyato2018spectral} to both the generator and discriminator as \cite{nazeri2019edgeconnect} to enhance the training stability. The TSMGAN loss $L_D$ for the discriminator to tell if the input video z is real or fake and $L_G$ for the generator to fool the discriminator are defined as:
\begin{equation}
\begin{aligned}
L_{D} &= \mathds{E}_{x\sim P_{data}(x)}[ReLU(1+D(x))] \\ &+
\mathds{E}_{z\sim P_{z}(z)}[ReLU(1-D(G(z)))]
\end{aligned}
\end{equation}

\begin{equation}
\begin{aligned}
L_{G} = -\mathds{E}_{z\sim P_{z}(z)}[D(G(z))]
\end{aligned}
\end{equation}

As \cite{yu2018free}, the kernel size is $5 \times 5$, stride $2 \times 2$ and the shifting operation is applied to all 11 convolutional layers for the TSMGAN discriminator, so the receptive field of each output feature point includes the whole video. It could be viewed as several GANs on different feature points, and a local-global discriminator structure \cite{yu2018generative} is thus not needed. 

Besides, the TSMGAN learns to classify real or fake for each spatial-temporal feature point from the last convolutional layer in the discriminator, which mostly consists of high-level features. Since the $l_1$ loss already focuses on low-level features, using TSMGAN could improve the model in an efficient way.

\vspace{-3mm}
\paragraph{\textbf{Overall loss.}}
The overall loss function to train the model is defined as:
\begin{equation}
\begin{aligned}
L_{total} &= \lambda_{l_1} L_{l_1}    +
\lambda_{perc}  L_{perc}         +
\lambda_{style} L_{style} + 
\lambda_{G} L_{G}
\end{aligned}
\end{equation}
where $\lambda_{l_1}$, $\lambda_{perc}$, $\lambda_{style}$ and $\lambda_{G}$ are the weights for $l_1$ loss, perceptual loss, style loss and TSMGAN loss, respectively.

\subsection{Network Design}
The model has a U-net like generator and a TSMGAN discriminator. The generator is composed of 11 convolutional layers with the proposed Gated Temporal Shift Module, including down-sampling, dilated and up-sampling ones. Similar structures are also adopted for state-of-the-art image inpainting models \cite{yu2018free,liu2018image}. Unlike U-net, there is no skip connection as there are many masked areas in the down-sampling layers. For down-sampling and up-sampling layers, we apply bilinear interpolation before convolutions.

\begin{table}[]
\begin{tabular}{|c|c|c|ccccc|}
\hline
Dataset       & Metric       & Mask   & TCCDS   & EC & CombCN & 3DGated & LGTSM \\ \hline
\multirow{9}{*}{\begin{tabular}[c]{@{}c@{}}FaceFo\\ -rensics\end{tabular}} & \multirow{3}{*}{MSE$\downarrow$}  
& Curve  & 0.0031* & 0.0022    & 0.0012 & \textbf{0.0008} & 0.0012\\
 &     & Object & 0.0096* & 0.0074    & \textbf{0.0047} & 0.0048  & 0.0053\\
& & BBox   & 0.0055  & 0.0019    & \textbf{0.0016} & 0.0018 & 0.0020 \\ \cline{2-8} 
 & \multirow{3}{*}{LPIPS$\downarrow$} & Curve  & 0.0566* & 0.0562    & 0.0483 & \textbf{0.0276} & 0.0352 \\
& & Object & 0.1240* & 0.0761    & 0.1353 & \textbf{0.0743} & 0.0770\\
& & BBox   & 0.1260  & \textbf{0.0335}    & 0.0708 & 0.0395 & 0.0432 \\ \cline{2-8} 
& \multirow{3}{*}{FID$\downarrow$}   & Curve  & 1.281*  & 0.848     & 0.704  & \textbf{0.472}  & 0.601\\
& & Object & 1.107*  & 0.946     & 0.913  & \textbf{0.766}  & 0.782 \\
& & BBox   & 1.013   & \textbf{0.663}     & 0.742  & \textbf{0.663}  & 0.681 \\ \hline
\multirow{6}{*}{FVI}   & \multirow{2}{*}{MSE$\downarrow$}   & Curve  &  0.0220* &  0.0048 &  \textbf{0.0021} &  0.0025 &  0.0028 \\
& & Object &  0.0110* &  0.0075 &  \textbf{0.0048} &  0.0056 &  0.0065 \\ \cline{2-8} 
& \multirow{2}{*}{LPIPS$\downarrow$} & Curve  &  0.2838* &  0.1204 &  0.0795 &  \textbf{0.0522} &  0.0569 \\
& & Object &  0.2595* &  0.1398 &  0.2054 &  \textbf{0.1078} &  0.1086 \\ \cline{2-8} 
& \multirow{2}{*}{FID$\downarrow$}   & Curve  &  2.1052* &  1.0334 &  0.7669 &  \textbf{0.6096} &  0.6436 \\
& & Object &  1.2979* &  1.0754 &  1.0908 &  \textbf{0.9050} &  0.9323 \\ \hline
\multicolumn{3}{|c|}{\begin{tabular}[c]{@{}c@{}}Number of Parameters$\downarrow$ \\ (Generator/Discriminator)\end{tabular}}  & -    & 20M/6M    & 16M/- & 36M/18M & \textbf{12M}/6M      \\ \hline
\multicolumn{3}{|c|}{Inference FPS$\uparrow$}  & 0.05$^\#$   & 55    & \textbf{120} & 23 & 80     \\ \hline
\end{tabular}
\caption{Quantitative comparison with baseline models on the FaceForensics and FVI dataset based on \cite{chang2019free}. The results of FVI dataset are averaged of seven mask-to-frame ratios; detailed results of each ratio could be found in the supplementary materials. *TCCDS failed on some cases and the results are averaged of successful ones. $^\#$runs on CPU; others are on GPU.}
\label{tab:quantitative_FVI}
\vspace{-4mm}
\end{table}
\section{Experimental Results}

\subsection{Setups}

\paragraph{Datasets.}
To compare with the baselines \cite{Huang-SigAsia-2016, wang2018videoinp, nazeri2019edgeconnect,chang2019free}, we train and test our model on the FaceForensics \cite{rossler2018faceforensics} and Free-form Video Inpainting (FVI) \cite{chang2019free} dataset. Both datasets are based on videos from YouTube, so they are close to real world scenarios. FaceForensics is composed of 1,004 videos with face, news-caster or newsprogram tags. There are only frontal faces cropped to $128 \times 128$ in the FaceForensics dataset, so it is rather simple for learning based models. Amongst, 150 videos are for evaluation while the rest are for training.

On the other hand, the FVI dataset contains 15,000 high-resolution videos with human activities, animals, natural scenes, etc. It also provides algorithms to generate free-form video masks for training. We re-size videos to $320 \times 180$ and split 100 videos for evaluation following the setup in \cite{chang2019free}. Note the FVI dataset is considered more challenging as the videos are very diverse.

\vspace{-3mm}
\paragraph{Training and Testing.}
Empirically we found that the model converges slower when directly trained as a whole. Thus, during training, we first pre-train the generator without the TSMGAN loss until convergence, and then fine-tune with the TSMGAN. We initialize the temporal shifting kernels in the LGTSM with the values that are equivalent to the original TSM. The pre-train stage takes about 1 day, while the fine-tune stage takes about 3 days on the FVI dataset, which is 3 times faster, reducing training time from 10 days to 3 days, than the model with 3D convolutions \cite{chang2019free}, demonstrating the merits of the proposed module. For other implementation details, please see the supplementary materials.

\begin{figure} \centering
\includegraphics[width=\linewidth]{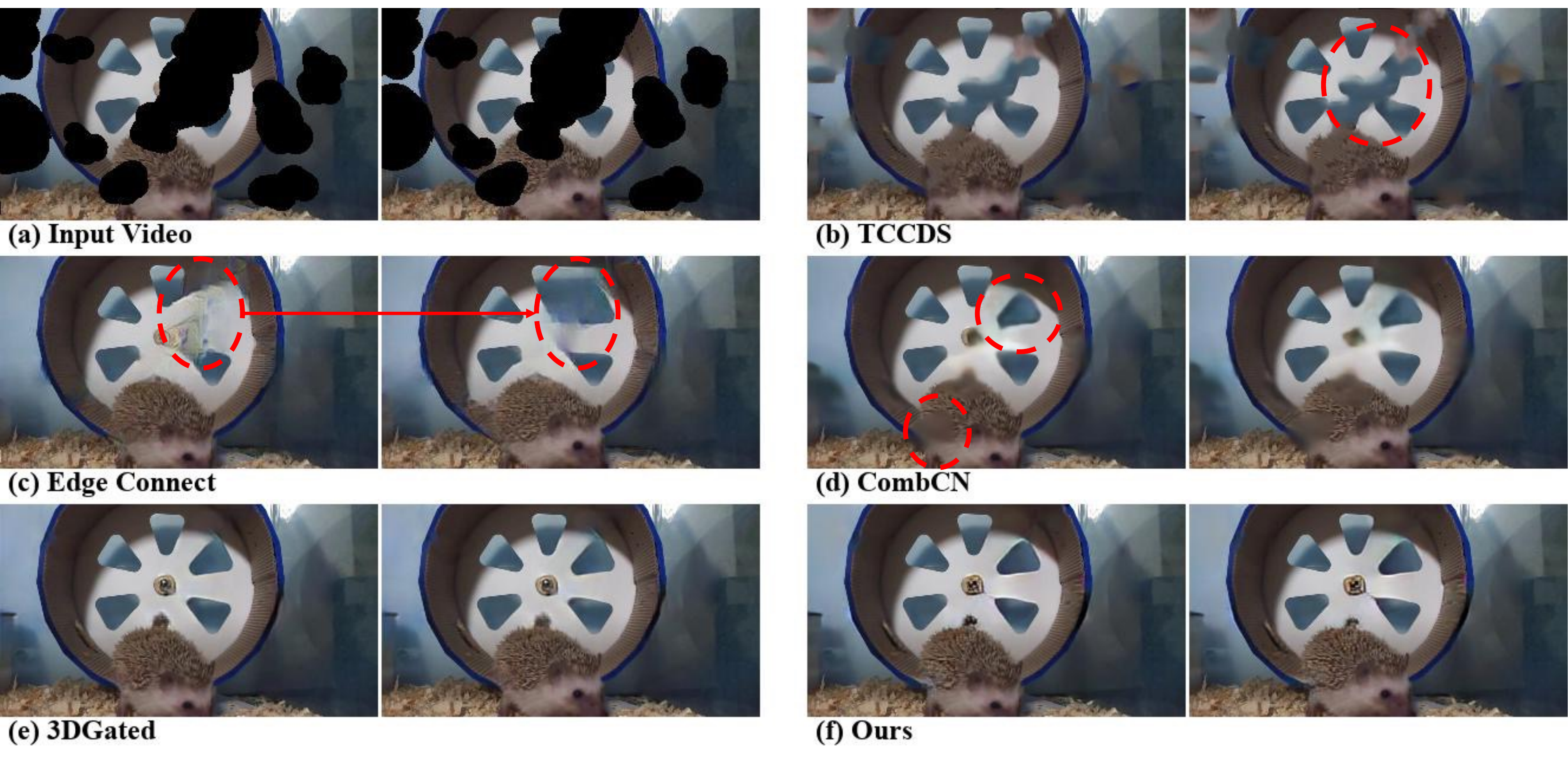}
\caption{Visual comparison with the baselines on the FVI testing set with object-like masks. Best viewed in color and zoom-in. See \href{https://www.youtube.com/watch?v=87Vh1HDBjD0&list=PLPoVtv-xp_dL5uckIzz1PKwNjg1yI0I94&index=32&t=0s}{video}.}
\label{fig:visual_comparison}
\vspace{-4mm}
\end{figure}

\vspace{-2mm}
\subsection{Quantitative Results}
As \cite{chang2019free}, mean square error (MSE) and Learned Perceptual Image Patch Similarity (LPIPS) \cite{zhang2018unreasonable} are used to evaluate the image quality; the Fr\'{e}chet Inception Distance (FID) \cite{heusel2017gans} with pre-trained I3D \cite{carreira2017quo} is used to evaluate video quality and temporal consistency.

We compare with state-of-the-art baselines with different strategies: the patch-based video inpainting method TCCDS by Huang \etal \cite{Huang-SigAsia-2016}, two-stage deep image inpainting method Edge-Connect (EC) by Nazeri \etal \cite{nazeri2019edgeconnect}, two-stage deep video inpainting method CombCN by Wang \etal \cite{wang2018videoinp}, and one-stage deep video inpainting method with 3D gated convolutions (3DGated) by Chang \etal \cite{chang2019free}. We train all learning based models on the FaceForensics and FVI datasets with free-form masks from \cite{chang2019free} for fair comparison. 

The averaged results of 7 ranges of mask-to-frame ratio from  0 - 10\% to 60 - 70\% on the FaceForensics and FVI testing are shown in Table \ref{tab:quantitative_FVI}. We could see that
our model is on par with the state-of-the-art method 3DGated \cite{chang2019free} in terms of perceptual distance (LPIPS and FID) and video quality (FID) with only 33\% of parameters and inference time (note that the results are averaged; our model performs better for some mask-to-frame ratios). TCCDS failed on many cases since the masks are irregular and it cannot properly recover partially masked objects. Edege-connect (EC) performs better on FaceForensics dataset with bounding box masks because faces are all aligned and the generated edges could be stable. Still, it has serious temporal inconsistent problem under other circumstances (see Fig. \ref{fig:visual_comparison}). Although CombCN has lowest MSE scores, it could only generates blurry results for the FVI dataset (see Fig. \ref{fig:visual_comparison}) and require more parameters for the generator than our model.

\vspace{-3mm}
\subsection{Qualitative Results}
From Fig. \ref{fig:visual_comparison} we could observe that the proposed model outperforms TCCDS (wrong patches), Edge-connect (temporally inconsistent) and CombCN (blurry), while almost the same as 3DGated. More visual comparison could be found in the supplementary materials.

\vspace{-3mm}
\subsection{Ablation Study}
In order to validate the contribution of each component, we also conduct an ablation study on the FVI dataset (Table \ref{tab:ablation_study}). We could observe that both the gated convolution and TSMGAN play important roles in our model; without them, the model will have a significant drop of performance. The proposed learnable shifting kernel further improves the performance with almost no additional parameters and achieve state-of-the-art results. Visual comparison be found in the supplementary materials and \href{https://www.youtube.com/watch?v=B8aCnWQw-9Y&list=PLPoVtv-xp_dLRHUthPXMhZX-O6IBq1RYV&index=14&t=0s}{videos}. 

\begin{table}
 \begin{center}
\begin{tabular}{|ccccc|ccc|}
\hline
\begin{tabular}[c]{@{}c@{}}3D \\ conv. \end{tabular} &
\begin{tabular}[c]{@{}c@{}}TSM\end{tabular} &
\begin{tabular}[c]{@{}c@{}}Learnable \\ shifting \end{tabular} &
\begin{tabular}[c]{@{}c@{}}Gated \\ conv. \end{tabular} & 
\begin{tabular}[c]{@{}c@{}c@{}}GAN  \end{tabular} & 

LPIPS$\downarrow$  & FID $\downarrow$ & Param. $\downarrow$ \\ \hline
$\boldcheckmark$& & &   $\boldcheckmark$ &   $\boldcheckmark$ &  0.1209  &  1.034 & 36M+18M \\  \hline
&$\boldcheckmark$  &  &   &   $\boldcheckmark$  &   0.2048   &    1.303 & 12M*+6M \\
&$\boldcheckmark$  &  &  $\boldcheckmark$ &      &   0.1660  &   1.198 & 12M  \\
&$\boldcheckmark$  &  & $\boldcheckmark$  & $\boldcheckmark$ &  0.1256  & 1.091 & 12M+6M\\
&$\boldcheckmark$ &$\boldcheckmark$ & $\boldcheckmark$  & $\boldcheckmark$ &  \textbf{0.1213}  & \textbf{1.039 } & 12M+6M\\
 \hline
\end{tabular}
\caption{Ablation study on the FVI dataset with object-like masks. Number of parameters are shown as generator/discriminator. *We increase the channel of vanilla convolution to fairly compare with gated convolutions.}
\label{tab:ablation_study}
\vspace{-5mm}
\end{center}
\end{table}

\vspace{-3mm}
\section{Discussion and Future Work}
Our model achieves state-of-the-art performance with LGTSM and 2D convolutions. However, the performance is still a bit lower than the 3D convolutions in \cite{chang2019free}. How to design a module that could handle temporal information in a more efficient way is still a challenging future work.

Another future work is to extend the input videos to higher or arbitrary resolution. For now, deep learning models are limited to a fixed resolution, which is not sufficient for exquisite videos. Simply applying these models to different parts of a high-resolution video may cause in spatial inconsistency.

\vspace{-3mm}
\section{Conclusion}
This paper presented a novel Learnable Gated Temporal Shift Module (LGTSM) for free-form video inpainting. LGTSM learns to shift some channels to its temporal neighbors in each frame and apply gating filter to attend on masked/inpainted/unmasked areas, which enables 2D convolutions to process temporal information and tackle poisoning masked areas at the same time.  In addition, LGTSM is highly efficient, using only 33\% of parameters and inference time compared to the state-of-the-art model with 3D convolutions. Experiments on the FaceForensics and FVI dataset suggest that the proposed model reach state-of-the-art performance in terms of evaluation metrics and visual results.

\section{Acknowledgement}
This work was supported in part by the Ministry of Science and Technology, Taiwan, under Grant MOST 108-2634-F-002-004. We also benefit from the NVIDIA grants and the DGX-1 AI Supercomputer. We are grateful to the National Center for High-performance Computing.

\bibliography{egbib}
\end{document}